\def\year{2022}\relax
\documentclass[letterpaper]{article} 
\usepackage{aaai22}  
\usepackage{times}  
\usepackage{helvet}  
\usepackage{courier}  
\usepackage[hyphens]{url}  
\usepackage{graphicx} 
\usepackage{natbib}  
\usepackage{caption} 
\DeclareCaptionStyle{ruled}{labelfont=normalfont,labelsep=colon,strut=off}

\usepackage{algorithm}
\usepackage{algorithmic}
\usepackage{times}
\usepackage{epsfig}
\usepackage{graphicx}
\usepackage{amsmath}
\usepackage{multirow}
\usepackage{helvet}
\usepackage{hyperref}
\usepackage{courier}
\usepackage{float}
\usepackage{color}
\usepackage{subfigure}
\usepackage{amssymb}
\usepackage{newfloat}
\usepackage{listings}
\floatstyle{ruled}
\newfloat{listing}{tb}{lst}{}
\floatname{listing}{Listing}
\makeatletter

\newcommand{\Rmnum}[1]{\expandafter\@slowromancap\romannumeral #1@}
\makeatother
\title{MTLDesc: Looking Wider to Describe Better}
\author{
Changwei Wang\textsuperscript{\rm 1,4,}\equalcontrib,
Rongtao Xu\textsuperscript{\rm 1,4,}\equalcontrib,
Yuyang Zhang\textsuperscript{\rm 1,4},
\\Shibiao Xu\textsuperscript{\rm 2,}\thanks{Shibiao Xu and Weiliang Meng are the corresponding authors (shibiaoxu@bupt.edu.cn, weiliang.meng@ia.ac.cn).},
Weiliang Meng\textsuperscript{\rm 1,3,4,\dag},
Bin Fan\textsuperscript{\rm 5},
Xiaopeng Zhang\textsuperscript{\rm 1,4}
}
\affiliations{
    \textsuperscript{\rm 1}NLPR, Institute of Automation, Chinese Academy of Sciences\\
    \textsuperscript{\rm 2}School of Artificial Intelligence, Beijing University of Posts and Telecommunications\\
        \textsuperscript{\rm 3}Zhejiang Lab;
    \textsuperscript{\rm 4}School of Artificial Intelligence, University of Chinese Academy of Sciences\\

    \textsuperscript{\rm 5}School of Automation and Electrical Engineering, University of Science and Technology Beijing}

\usepackage{bibentry}

\begin{document}

\maketitle

\begin{abstract}
Limited by the locality of convolutional neural networks, most existing local features description methods only learn local descriptors with local information and lack awareness of global and surrounding spatial context. In this work, we focus on making local descriptors ``look wider to describe better'' by learning local \textbf{Desc}riptors with \textbf{M}ore \textbf{T}han just \textbf{L}ocal information (\textbf{MTLDesc}).
Specifically, we resort to context augmentation and spatial attention mechanisms to make our MTLDesc obtain non-local awareness.
First, Adaptive Global Context Augmented Module and Diverse Local Context Augmented Module are proposed to construct robust local descriptors with context information from global to local.
Second, Consistent Attention Weighted Triplet Loss is designed to integrate spatial attention awareness into both optimization and matching stages of local
descriptors learning.
Third, Local Features Detection with Feature Pyramid is given to obtain more stable and accurate keypoints localization.
With the above innovations, the performance of our MTLDesc significantly surpasses the prior state-of-the-art local descriptors on HPatches, Aachen Day-Night localization and InLoc indoor localization benchmarks.\href{https://github.com/vignywang/MTLDesc}{[Code release]} 
\end{abstract}

\section{Introduction}
Local descriptors currently play a key role in various vision applications such as image matching, image retrieval, SfM, SLAM, and visual localization.
With the industry's rapid development, these applications must deal with more complex and challenging scenarios (various conditions such as day, night, and seasons).
As the local features detection and description are the critical for these applications, there is an urgent need to further boost their performance.

\begin{figure}[t]
\begin{center}

  \includegraphics[width=1 \linewidth]{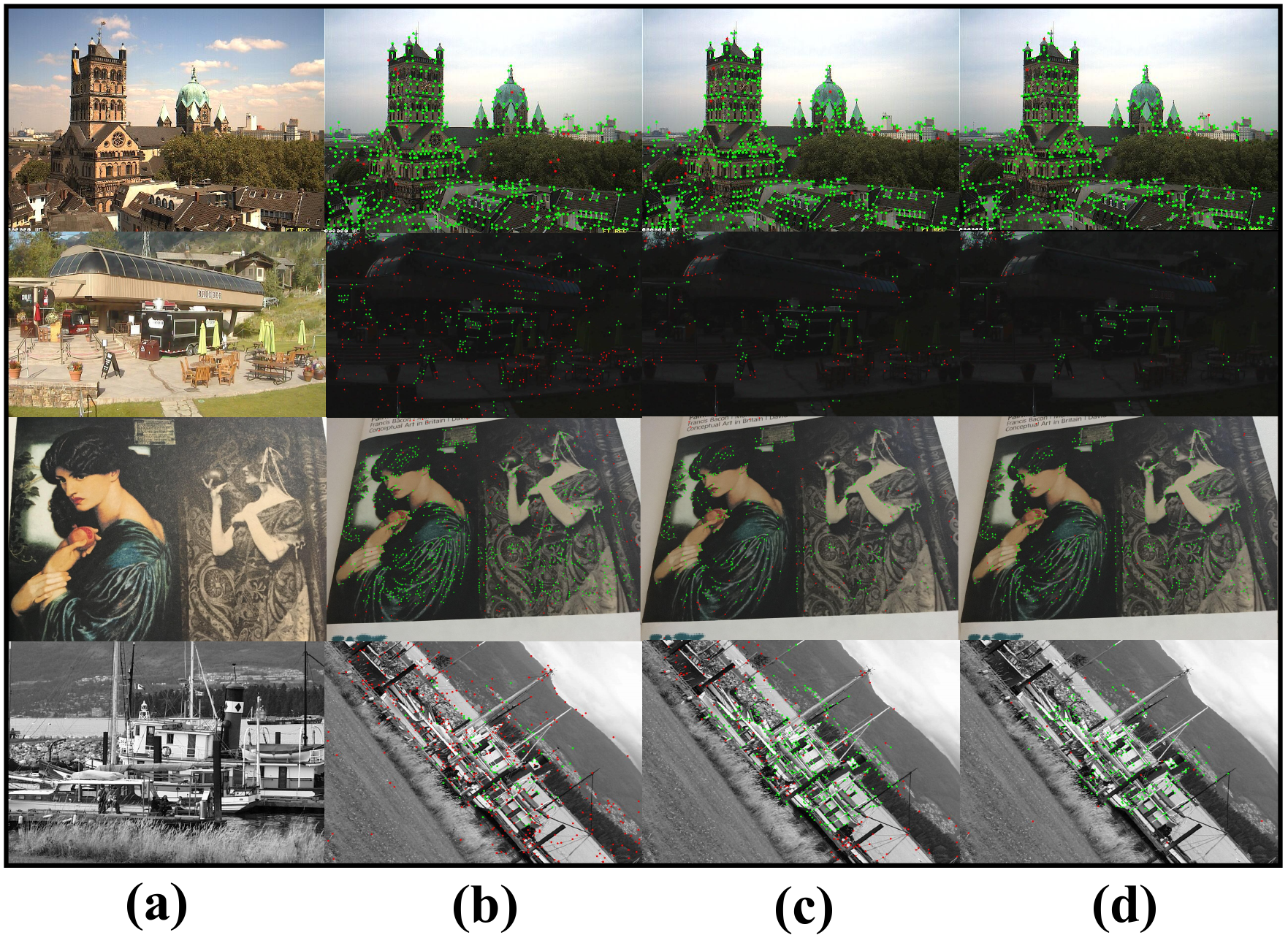}   
\end{center}
   \caption{
   Matching results under illumination and viewpoint changes. {\bf (a)}: Matching images. {\bf (b)}: Baseline (SuperPoint). {\bf (c)}: MTLDesc w/ Context Augmentation but w/o Consistent Attention Weighting. {\bf (d)}: MTLDesc both w/ Context Augmentation and w/ Consistent Attention Weighting. {\color{green}{\bf Green dots}}: Correct matches. {\color{red}{\bf Red dots}}: Incorrect matches.
\vspace{-0.cm}
  }
\label{fig:introduction}
\end{figure}

Geoffrey Hinton said that “local ambiguities have to be resolved by finding the best global interpretation” in his first paper~\cite{hinton1976using}. This idea still holds true in local descriptors learning.
There are two main weaknesses for learning local descriptors only using limited local visual information:
i) Ambiguity regions with repetitive patterns (texture, color, shape, etc.) is difficult to be distinguished only by local information, as shown in the first row (trees) and the fourth row (river and ground) in Fig.~\ref{fig:introduction} {\bf (b)};
ii) Local visual information becomes unreliable and indistinguishable for challenging scenes with large illumination and viewpoint changes, which will lead to massive incorrect matches (the second and fourth rows of Fig.~\ref{fig:introduction} {\bf (b)}). 
On this account, robust non-local context can be employed to better distinguish challenging local regions. 
We propose Context Augmentation and Consistent Attention Weighting to look wider for describing better, enabling our descriptors to gain awareness beyond the local region, in turn to effectively mitigate above weaknesses as shown in Fig.~\ref{fig:introduction} {\bf (d)}.



CNN-based backbones like L2Net~\cite{l2net} and VGG~\cite{vgg} are widely adopted by the local descriptors learning methods. Due to the inherent locality of CNN, features can only be extracted in a limited receptive field. 
Although some of these methods~\cite{luo2019contextdesc,aslfeat,tyszkiewicz2020disk} can implicitly alleviate this problem by extracting features in a larger receptive field, they still only considered the context in a fixed patch-wise neighborhood.
By contrast, our method further utilizes the context of the global and different receptive fields by the well-designed Adaptive Global Context Augmented Module (\textbf{AGCA}) and Diverse Local Context Augmented Module (\textbf{DLCA}).

In addition, the attention mechanism is also an effective way to use non-local information. When humans observe and describe images, they will quickly analyze spatial information and focus their attention on some key regions. Some image retrieval practices~\cite{DELG} have indicated that the network can also obtain similar awareness with the attention mechanism. 
Based on the above inspiration, we design a new Consistent Attention Mechanism for both optimization and matching of local descriptors by a well-designed Consistent Attention Weighted Triplet Loss.

Furthermore, as the localization accuracy of keypoints also affects the results of local features matching, we propose a Local Features Detection with Feature Pyramid based on the classical scale-space~\cite{lowe2004distinctive} and deep supervision~\cite{lee2015deeply} to obtain more stable and accurate keypoints localization.
Besides, to meet the demand of practical applications, running speed should also be addressed. Therefore, the whole method is carefully designed to be as fast as possible.
In summary, there are four main contributions in this paper:~\textbf{1)} We devise AGCA and DLCA modules to aggregate effectively from global to local context information for local descriptors learning.
\textbf{2)} We propose a novel Consistent Attention Weighted Triplet Loss to introduce spatial consistent attention awareness in both optimization and matching of local descriptors. \textbf{3)} We present the Local Features Detection with Feature Pyramid to improve the localization accuracy of keypoints. \textbf{4)} We provide a \textbf{real-time} solution for local features learning named MTLDesc, which achieves state-of-the-art on HPatches, Aachen-Day-Night and InLoc benchmarks.

\section{Related Works}
\begin{figure*}[htbp]
\begin{center}
  \includegraphics[width=1 \linewidth]{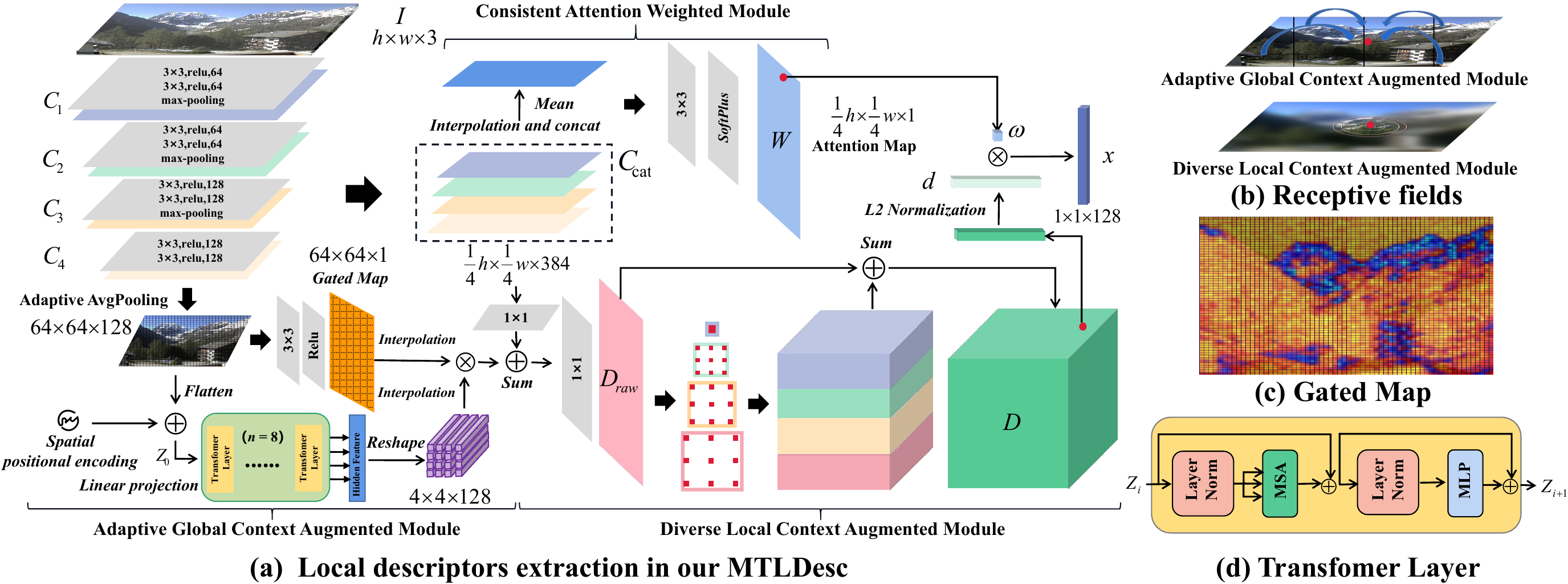}
\end{center}
   \caption{(a): Local descriptors extraction in our MTLDesc.
   (b): Receptive fields of proposed modules in (a). (c): Gated Map in (a). The values of the {\color{blue} {\bf blue regions}} are 0 and the others are positive values. (d): Details of Transfomer Layer in (a). 
\vspace{-0.5cm}
}
\label{fig:descriptor}
\end{figure*}

\noindent {\bf Local Descriptors Learning:}
Hand-crafted local descriptors are widely used in computer vision, and are comprehensively evaluated in ~\cite{traditional_evaluation2}. 
Current deep learning based descriptors can roughly fall into two categories: patch-based descriptors and dense descriptors.
Patch-based descriptors~\cite{l2net,hardnet,sosnet,luo2019contextdesc} extract descriptors from corresponding patches of the detected keypoints (e.g. SIFT~\cite{sift}), while dense descriptors~\cite{superpoint,r2d2,d2net,r2d2,aslfeat,wang2020caps,tyszkiewicz2020disk} usually use a fully convolutional neural network ~\cite{fcn} to extract dense feature descriptors for the whole image in one forward pass. Dense descriptors have achieved good performance in image matching and long-term visual localization, showing great potential for practical applications. 
In contrast to these prior works, we recommend the introduction of non-local information include both context and spatial attention awareness to local descriptors learning, aiming to make the local descriptors ``look wider to describe better''.

\textbf{Context Awareness:}
Context awareness is essential for pixel-level tasks~\cite{bisenet}, but it has not been attracted widespread attention in local descriptors learning.
ContextDesc~\cite{luo2019contextdesc} aggregates the cross-modality context for local descriptors, including visual context from a ResNet-50~\cite{resnet} branch, and geometric context from keypoints distribution.
Patch-based ContextDesc depends on a large network and needs to obtain the geometric context after additional keypoints detection calculation, so it consumes more computing time and memory space.
Some CNN-based dense descriptors implicitly improve context awareness by increasing the receptive field. ASLFeat~\cite{aslfeat} proposes to use deformable convolution~\cite{dai2017deformable} to extract descriptors with shape context,
MLIFeat~\cite{mlifeat} utilizes hypercolumns~\cite{hariharan2016object} to fuse multi-level features, while DISK~\cite{tyszkiewicz2020disk} employs UNet-like backbone~\cite{unet} to fuse multi-scale context information. 
However, all these descriptors only aggregate the context of a fixed receptive field and do not consider the global context. 
Recently, Visual Transformers~\cite{vit} has shown the ability to aggregate global context in some computer vision applications~\cite{transformersdet,transformersseg}. 
In this work, we make the network get comprehensive context awareness from global to local.
On the one hand,  we use visual transformer and a learnable Gated Map to adaptively embed the global context and location information into local descriptors. 
On the other hand, we propose to flexibly learn local descriptors through surrounding contexts with different receptive fields.



\textbf{Attention Mechanism:} As a non-local awareness, spatial attention has been successfully applied to the learning of image-level global descriptors~\cite{attention2,DELF,DELG,attention1} for image retrieval.
In these methods, spatial attention is used as the weight of local descriptors, and global descriptors are derived from local descriptors through weighted summation. 
However, directly applying the local descriptors of these methods to image matching produces poor results, as reported by~\cite{r2d2,d2net}.
This may be caused by the lack of supervision of local pixel correspondence. However, attention mechanisms in these methods are optimized with the supervision of image-level and it is not suitable for pixel correspondence supervision. 
In contrast, we proposes a special consistent attention mechanism to improve the optimization and matching of local descriptors for image matching.
   
\section{Method}
Our MTLDesc employs a fully convolutional network encoder as the \textbf{shared backbone} for both local features description and detection.
The encoder consists of $3 \times 3$ convolutional layers, relu layers, and max-pooling layers. For a $h \times w$ image $I$, $C_{1}(h\times w)$ ,$ C_{2}(h/2\times w/2)$, $C_{3}(h/4\times w/4)$, and $C_{4}(h/8\times w/8)$ feature maps are obtained after four sequential sub-encoders. 
MTLDesc detects keypoints (\textbf{Fig.~\ref{fig:detector}}) and extracts corresponding local descriptors (\textbf{Fig.~\ref{fig:descriptor}}) at the same time, and the two parts use the same shared backbone network.
\vspace{-0.1cm}
\subsection{Local Descriptors with Non-local Information}
\label{sec:Local Descriptor}
%
\subsubsection{\Rmnum{1}. Local Descriptors with Context Augmentation:}
We propose Adaptive Global Context Augmented Module and Diverse Local Context Augmented Module to implement context augmentation from global to local, as shown in Fig.~\ref{fig:descriptor}~(a), while Fig.~\ref{fig:descriptor}~(b) shows the difference between the two modules about the receptive field, which leads them to aggregate context from different perspectives.

\vspace{0.15cm}
\noindent\textbf{ (1) Adaptive Global Context Augmented Module.} Different image regions usually have different degrees of demand for global context. For regions that are difficult to describe only with local information (e.g. weak or repeated textures regions), the global context can introduce more spatial information to make it more discriminative. But for regions with good distinguishable local information, directly adding the global context may bring some noise. 
Our Adaptive Global Context Augmented Module is designed to adaptively introduce global context for local descriptors.
Specifically, we take the feature maps $C_{4}$ through adaptive average pooling to obtain a fixed-size feature map ($64\times 64$) as the input of the module. Compared with previous visual transformers, our method can effectively reduce the computational complexity and adapt to images of any size. 
Following~\cite{vit}, we perform tokenization by reshaping the input into a sequence of flattened 2D patches $X_{p}$, and each patch is of size $16\times 16$.
We map the vectorized patches $X_{p}$ a latent 128-dimensional embedding space using a trainable linear projection. In order to make the local descriptors obtain the spatial position information relative to the global, we learn specific position embeddings which are added to the patch embeddings to retain positional information as follows: $Z_{0}=[X_{p}^{1}E;X_{p}^{2}E;...;X_{p}^{N}E]+E_{pos}$, where $E$ is the patch embedding projection and $E_{pos}$ is the position embedding. After $Z_{0}$ passes through 8 transformer layers, the hidden features with global context are obtained.
The structure of transformer layer is shown in Fig.~\ref{fig:descriptor}~(d). Where MSA denotes Multihead Self-Attention and MLP denotes Multi-Layer Perceptron block~\cite{vit}. After reshaping, we can get the patch-wise descriptors with the global context.
Another branch of the module predicts a gated map to mask regions that do not require a global context supplement. We implement the gating mechanism through the relu activation function, and the visualization result of gated map is shown in Fig.~\ref{fig:descriptor}~(c). Finally, we merge the global context filtered by the gated map into local descriptors.

\noindent\textbf {(2) Diverse Local Context Augmented Module.}
The surrounding context is also crucial for local descriptors learning. We design a simple and effective  Diverse Local Context Augmented Module to extract diverse surrounding contexts.
Unlike most previous CNN-based local descriptors which only deploy the top-layer feature maps to extract descriptors, we recommend using all feature maps derived from the backbone to construct descriptors. Specifically, the $C_{1}, C_{2}, C_{3},C_{4}$ are interpolated to the same spatial size and then aggregated together to get $C_{cat}$. The size is set to $1/4$ of the input image $I$ as it gives a good trade-off between accuracy and speed. It improves the utilization of feature maps and integrates information of different scales.
To obtain diverse surrounding contexts, we decouple the descriptor into four 32-dimensional sub-descriptors and learn them in different description spaces respectively. This ensures that the sub-descriptors remain independent and diversity to contain more information.
Specifically, we use $1\times1$ Conv and three $3\times3$ dilated Conv~\cite{dilated} with dilation rates of 6, 12 and 18 respectively to derive descriptors from different receptive fields. After being concatenated and added to $D_{raw}$, the final dense descriptor $D$ is obtained. 
The surrounding context from different receptive fields further stimulates the representation ability of local descriptors.

\begin{figure*}[htb]
\begin{center}
\setlength{\abovecaptionskip}{-0.5 cm}
  \includegraphics[width=1\linewidth]{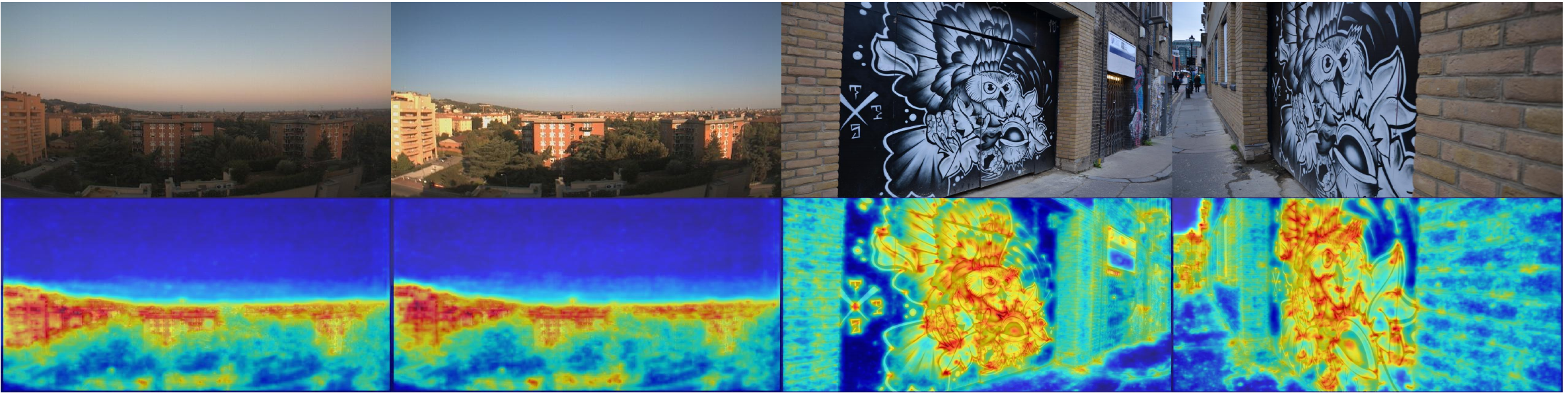}
\end{center}
   \caption{
  Images (first row) and {\bf Consistent Attention Maps} (second row) under illumination or viewpoint changes. For Consistent Attention Maps, pixels close to {\color{red}{\bf red}} means higher attention score and close to  {\color{blue}{\bf blue}} means lower score. Meaningless regions (e.g. sky and ground) and regions with repetitive texture (e.g. trees and brick wall) are given low attention scores.
  \vspace{-0.5cm}
   }
\label{fig:attention}
\end{figure*}

\subsubsection{\Rmnum{2}. Local Descriptors with Consistent Attention Weighting:}
To further overcome the limitation of keypoints description only based on local information, we make the network imitate humans to obtain the awareness and insight of spatial information with a consistent attention mechanism.
We claim that the proposed Consistent Attention has \textbf{three types of properties}:
 \textbf{\textit{i)}} The same regions in different images have consistent attention scores. 
\textbf{\textit{ii)}} Representative regions are given higher attention scores, as these regions easily match to inliers while distinguishable to outliers.
\textbf{\textit{iii)}} Descriptors from regions with high attention scores are optimized first.
We now explain how we design the module and loss for applying consistent attention to make the optimization and matching of local descriptors better. 

\noindent{\bf Consistent Attention Weighting Module} is shown in Fig~\ref{fig:descriptor}~(a).
Specifically, $C_{cat}$ is averaged across the channel dimension, and the attention map $W$ is predicted from this averaged feature map via the $3 \times 3$ Conv + SoftPlus.

\noindent{\bf Consistent Attention weighted Triplet Loss} is designed to jointly optimize local descriptors and consistent attention. Considering an image pair~$(I,I^{\prime})$, the dense descriptors $D,D^{\prime}$ and attention maps $W,W^{\prime}$ of $I,I^{\prime}$ can be extracted by our MTLDesc, as shown in Fig.~\ref{fig:descriptor}~(a). Given the sampled points set $P$ of size $N$ in $I$ and their corresponding points $P^{\prime}$ in $I^{\prime}$ , the corresponding descriptors of $P,P'$ are denoted as $d_{i}$ and ${d^{\prime}_{i}, i \in 1\dots N}$.
The corresponding score of the descriptor $d_{i}$ on the attention map $W$ is $\omega_{i}$, so the attention weighted descriptor is defined as $x_{i}=\omega_{i}\cdot d_{i}$ .
For $x_{i}$, its \textit{positive distance} ${||x_{i}||}^{+}$ is defined as:
{\setlength\abovedisplayskip{1pt}
\setlength\belowdisplayskip{1pt}
\begin{equation}
    {||x_{i}||}^{+}=||\omega_{i}\cdot d_{i} - \omega^{\prime}_{i}\cdot d^{\prime}_{i}||_2,
    \label{eq:positive_pair}
\end{equation}}
and its \textit{hardest negative distance} ${||x_{i}||}^{-}$ is defined as:
{\setlength\abovedisplayskip{1pt}
\setlength\belowdisplayskip{1pt}
\begin{equation}
    {||x_{i}||}^{-}=\underset{j \in 1\dots N, j\not=i}{\rm{min}}(||\omega_{i}\cdot d_{i} - \omega^{\prime}_{j}\cdot d^{\prime}_{j}||_2).
\label{eq:negative_pair}
\end{equation}}
The Consistent Attention Weighted Triplet Loss $\mathcal L_{\rm{Atrip}}$ can be defined as:
{\setlength\abovedisplayskip{1pt}
\setlength\belowdisplayskip{1pt}
\begin{equation}
    \mathcal L_{\rm{Atrip}}(x) =\frac{e^{^{\omega/T}}}{\sum_{i=1}^{N} e^{^{\omega_{i}/T}}} \max(0,{||x||}^{+}-{||x||}^{-}+1),
    \label{eq:attention triplet}
\end{equation}}
where $\omega$ is the attention score corresponding to $x$, and $T$ is a smoothing factor. $T$ is used to adjust the effect of attention weight on loss. 
We will explore the impact of $T$ in next section.
The whole description loss is summed as:
{\setlength\abovedisplayskip{1pt}
\setlength\belowdisplayskip{1pt}
\begin{equation}
 \mathcal{L}_{\rm{des}} = \sum_{i=1}^{N} \mathcal {L}_{\rm{Atrip}} (x_{i}),
    \label{eq:whole_triplet}
\end{equation}}
\vspace{-0.5cm} 
\begin{figure}[H]
\begin{center}
  \includegraphics[width=0.9\linewidth]{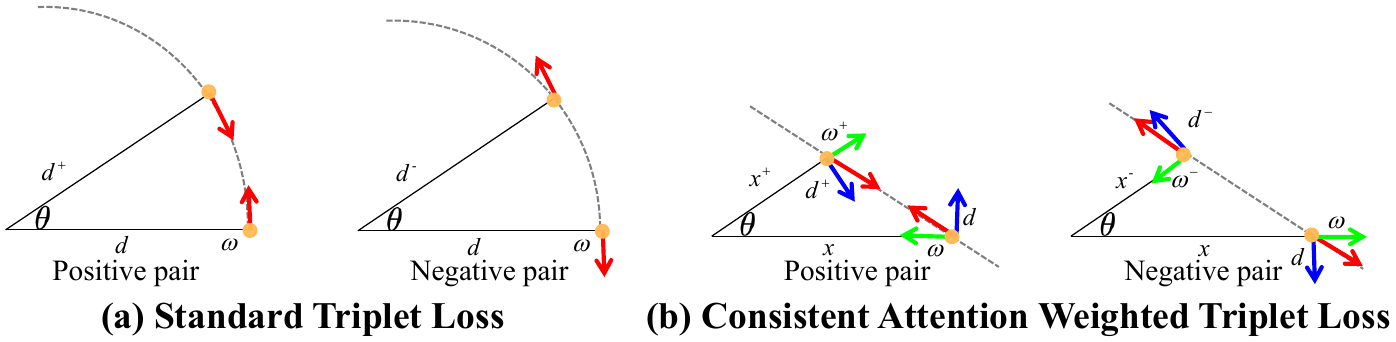}
\end{center}
   \caption{
   Example of the optimization direction of 2D descriptor.
  {\color{red}{\textbf{Red}}} arrow: Gradient descent direction.  
  {\color{green}{\textbf{Green}}} arrow: Gradient component of consistent attention $\omega$ optimization. 
  {\color{blue}{\textbf{Blue}}} arrows: Gradient component of  descriptor $d$ optimization.}
\label{fig:optimization}
\end{figure}
\label{paper:optimization}
\vspace{-0.3cm} 
\noindent{\bf (1) Consistent Attention in Optimization:}
First, we will discuss the difference between our $\mathcal L_{\rm{Atrip}}$ and standard triplet loss in the optimization direction. 
It is easy to validate the optimization of the L2 normalized descriptor's distance degenerates to angle's optimization~\cite{tian2020hynet}, meaning that the common standard triplet loss only has the optimization of the angle component as shown in Fig.~\ref{fig:optimization} (a). However, the optimization direction of our $\mathcal L_{\rm{Atrip}}$ is decoupled into the component of the descriptor $d$ (angle) optimization and the component of the consistent attention $\omega$ (weight) optimization in Fig.~\ref{fig:optimization} (b). For descriptors, $\mathcal L_{\rm{Atrip}}$ still optimizes the angle between them. For consistent attention, as shown in Fig.~\ref{fig:optimization}~(b), the attention scores of positive samples tend to convergent while the attention scores of negative samples tend to divergent. This trend will lead to the consistent distribution of attention scores ({i.e.,} {\bf property \textit{i)}}) in corresponding regions of image pairs. 
The details are shown in Fig.~\ref{fig:attention}.

Second, we will further explore the optimization goal of consistent attention based on the above discussion. As shown in Eq.~\ref{eq:attention triplet},
the consistent attention $\omega$ is affected by both triplet loss term $\max(0,{||x||}^{+}-{||x||}^{-}+1)$ which  only provides consistency as mentioned before and softmax term $\frac{e^{^{\omega/T}}}{\sum_{i=1}^{N} e^{^{\omega_{i}/T}}}$ in optimization.
To minimize the loss, the softmax term causes the network tends to give larger attention $\omega$ to samples with smaller triplet loss term. These samples usually have large ${||x||}^{-}$ and small ${||x||}^{+}$, so they are more suitable for matching. To minimize the total loss, these two conditions need to be met together, meaning that consistent attention has \textbf{properties} {\bf \textit{i)}} and {\bf\textit{ii)}} as shown in Fig.~\ref{fig:attention}. 


Third, we will explore the role of attention score $\omega$ in the optimization of descriptor $d$ by analyzing the gradient.
Given weighted descriptor $x = \omega \cdot d$ and positive sample $x^{+}$, we can get the gradient of the positive distance to the descriptor $d$  by partial derivative as following:
{\setlength\abovedisplayskip{1pt}
\setlength\belowdisplayskip{1pt}
\begin{equation}
\frac{\partial ||x - x^{+}||_2}{\partial d} = 
\omega \cdot \frac{ x-x^{+}}{||x-x^{+}||_2}.
    \label{eq:dao1}
\end{equation}}
In Eq.~\ref{eq:dao1}, it is obvious that the gradient of descriptors is weighted by the attention score.
Moreover, as shown in Eq.~\ref{eq:attention triplet}, the gradient of $d$ is also weighted by the softmax term in the whole loss. Note that softmax term is also proportional to $w$, so in summary, the sample with high attention score will contribute more gradients relatively (\textit{i.e.,} \textbf{property} {\bf \textit{iii)}}).
 Obviously, not every sample’s descriptor is worthy of equal optimization. Forcing learning descriptions on pixels that are not suitable as descriptors (e.g. sky, grass and waves) 
 will bring noise and lead to sub-optimal results. Therefore, the local descriptor can be optimized more flexibly and selectively with the help of consistent attention weighting.

\noindent{\bf (2) Consistent Attention in Matching:}
\label{paper:Matching}
Interestingly, we find that attention-weighted local descriptors are also more suitable for matching.
As shown in Fig.~\ref{fig:attention}, the corresponding regions have similar consistent attention scores in different images, so consistent attention can be used as a prior information in the local descriptors matching. 
It is evident that regions with high attention scores are more likely to be successfully matched with the same high scores regions in another image. Thus, the weighted descriptor has a smaller matching space and lead to higher matching accuracy.


\vspace{-0.3cm}
\begin{figure}[H]
\begin{center}
  \includegraphics[width=1\linewidth]{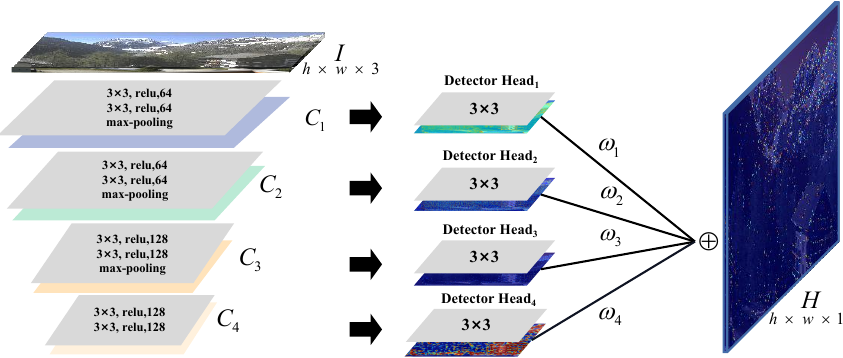}
\end{center}
   \caption{
  Local Features Detection with Feature Pyramid.
  \vspace{-0.3cm}
   }
\label{fig:detector}
\end{figure}

\subsection{Local Features Detection with Feature Pyramid}
\label{sec:Local detection}
We adopt pixel-wise classification to train keypoints detector with pseudo-keypoints supervision. Different from SuperPoint, we recommend predicting keypoints at different scale spaces by proposed Local Features Detection with Feature Pyramid. 
Specifically, we set four detection headers to predict the keypoint heatmaps respectively as shown in Fig~\ref{fig:detector}.
In order to combine the predict results,  we interpolate the predicted heatmaps of different scales to the image size $h\times w$. We set four learnable weights to fuse heatmaps of different scales to predict final keypoints and calculate the loss. 
Each detection header can receive direct supervision of detector loss, which can be considered as deep supervision~\cite{lee2015deeply}. Our method also conforms to the famous scale-space theory~\cite{lowe2004distinctive} which is accepted by many methods~\cite{keynet,aslfeat}, with the difference from them that we directly predict keypoints through supervised learning without additional statistics and calculations.
We use the weighted binary cross-entropy loss as the detector loss since there is an extreme imbalance in the number of keypoints and non-keypoints.
Given the predicted keypoints heatmap $K\in\mathbb{R}^{h\times w}$ and pseudo-ground truth label $G\in\mathbb{R}^{h\times w}$, the detector loss is defined as:
{
\setlength\belowdisplayskip{1pt}
\begin{equation}
\begin{aligned}
\mathcal L_{\rm{bce}}(k, g)=-\lambda g log(k)  -(1-g)log(1-k),
\end{aligned}
\label{eq:detection}
\end{equation}
\begin{equation}
\begin{aligned}
\mathcal L_{\rm{det}} = & \frac{1}{hw}\sum_{u,v}^{h,w}\mathcal L_{\rm{bce}}(K_{u,v}, G_{u,v})
\end{aligned}
\label{eq:bce}
\end{equation}}
where the weight $\lambda$ is empirically set to $200$.

\subsection{Training Strategy and Implementation Details}

\subsubsection{Data Preparation:}\label{sec:data}
We use MegaDepth~\cite{megadepth} to generate the training data with dense pixel-wise correspondences.
MegaDepth dataset contains image pairs with known pose and depth information from $196$ different scenes. 
Following the settings in D2-Net, we take 118 scenes from all scenes as the training set. 
We randomly select 100 image pairs from each scene, and intercept $400\times400$ image blocks from the original images for the training. 
Thus, we get $11,800$ image pairs with dense pixel correspondence.
This part of the data contains real complex transformations, which are difficult to collect but closer to practical applications.
Besides, we use random homography to synthesize more diverse image pairs to supplement the training data inspired by SuperPoint, further enriching the whole transformation types.
In summary, our training data consists of $23,600$ image pairs in total. We compare our dataset settings with advanced methods in the appendix.




\noindent{\bf Keypoints Supervision with Distillation:}
We employ distillation to get the pseudo-keypoints ground truth directly from an off-the-shelf trained SuperPoint (teacher model).
To obtain reliable and more pseudo-labeling of keypoints, we use iterative homographic adaptation to obtain the probability map of keypoints heatmap. We refer the readers to the work~\cite{superpoint} for more details. 

\noindent{\bf Correspondences Supervision with Keypoints Heatmap Guidance:}
Obviously, not all locations are equally important. 
Forcing the network to train the descriptors in meaningless areas may lead to impaired performance. 
To get enough keypoints-specific and distributed diversely correspondences, we design a novel keypoints guided correspondences sampling method.
Specifically, for each image pair $I_{1}, I_{2}$:
1) Predict keypoints heatmaps $M_1,M_2$ with a trained SuperPoint model.
2) Synthesize $M_1^{\prime}$ from $M_2$ based on the transformation between the image pair $I_{1}, I_{2}$.
3) Generate a compound keypoints heatmap by $M=M_1+M_1^{\prime}$ and divide this heatmap $M$ into $40\times40$ grids.
4) Take the point with the largest score of each grid on $M$ to obtain a candidate point set $Q$.
5) Apply the non-maximum suppression (NMS) to $Q$ and select the top 400 points to construct the refined descriptor correspondences $P$. 
\subsubsection{Implementations:}
To optimize the keypoints detection and description jointly, the total loss is composed of detector loss $\mathcal L_{det}$ and descriptor loss $\mathcal L_{des}$, which is formulated as:
{\setlength\abovedisplayskip{1pt}
\setlength\belowdisplayskip{1pt}
\begin{equation}
\mathcal L_{\rm{total}} = \mathcal L_{\rm{det}} + \mathcal L_{\rm{des}}.
    \label{eq:total_loss}
\end{equation}}
The Adam optimizer with poly learning rate policy is used to optimize the network, and the learning rate decays from $0.001$. 
The training image size is set to $400\times400$ with the training batch size $12$. 
The whole training process typically converges in 30 epochs and takes about 14 hours with a single NVIDIA Titan V GPU.
During the testing, the detection threshold $\alpha$ is empirically set to 0.9 and non-maximum suppression (NMS) radius to 4 to balance the keypoints' number and reliability.
Our method implemented by Pytorch\footnote{We also implemented our method by using Mindspore (https://www.mindspore.cn/) and observed similar performance.} runs at {\bf 24 FPS} (real time) on $480 \times 640$ images with a single NVIDIA Titan V GPU. 

\vspace{-0.1cm}
\section{Experiments}
\subsection{Comparisons on Image Matching}

\noindent\textbf{Dataset and Metrics:}
We use the popular HPatches benchmark~\cite{hpatches} for ablation studies and comparisons. 
Following previous methods, we use 108 sequences with viewpoint or illumination variations after excluding high-resolution sequences from 116 available sequences. 
The entire benchmark includes 56 sequences with changes in viewpoint and 52 sequences with changes in illumination.
We use three standard metrics for evaluation:
1) Mean matching accuracy (\textit{MMA}) is the average percentage of correct matches in image pairs under different matching error thresholds.
2) Match score (\textit{M.S.}) is the ratio of the correct match to the total number of keypoints estimated in the shared view, following the definition in~\cite{r2d2}.
3) Accuracy of homography (\textit{HA}) (~\cite{superpoint}) is used to compare the estimated homography of image pairs with its corresponding ground truth. 

\begin{table}[thb]
\centering
\resizebox{\linewidth}{!}{
\renewcommand\arraystretch{1.2}     {
\begin{tabular}{ccccc}
\hline
\multicolumn{5}{c}{HPatches dataset @3}                                                                                                                                                                                                                                                            \\ \hline
\multicolumn{1}{c|}{\textbf{Method}}                                                                                                            & \multicolumn{1}{c|}{\textbf{Config}}                                                          & \textbf{MMA\%} & \textbf{M.S.\%}  & \textbf{HA\%}  \\ \hline
\multicolumn{1}{c|}{\multirow{3}{*}{\begin{tabular}[c]{@{}c@{}}Baseline\\ (Improved SuperPoint)\end{tabular}}}                                           & \multicolumn{1}{c|}{\textit{SuperPoint orig.}}                                                & 64.44          & 42.41          & 72.59          \\
\multicolumn{1}{c|}{}                                                                                                                           & \multicolumn{1}{c|}{\textit{SuperPoint our impl.}}                                            & 67.51          & 43.54          & 71.48          \\ \cline{2-5} 
\multicolumn{1}{c|}{}                                                                                                                           & \multicolumn{1}{c|}{\textit{+ FP KeyPoints}}                                                  & \textbf{69.88} & \textbf{45.68} & \textbf{73.01} \\ \hline
\multicolumn{1}{c|}{\multirow{4}{*}{\begin{tabular}[c]{@{}c@{}}Baseline + Related \\ Comparison Methods\end{tabular}}}                                   & \multicolumn{1}{c|}{\textit{+ DCN}}                                                           & 70.56          & 44.19          & 72.14          \\
\multicolumn{1}{c|}{}                                                                                                                           & \multicolumn{1}{c|}{\textit{+ Hypercolumns}}                                                    & 70.35          & 45.89          & 73.56          \\
\multicolumn{1}{c|}{}                                                                                                                           & \multicolumn{1}{c|}{\textit{+ ASPP}}                                                          & 71.03          & 46.28          & 73.89          \\
\multicolumn{1}{c|}{}                                                                                                                           & \multicolumn{1}{c|}{\textit{+ UNet}}                                                                   & 71.15          & 45.78          & 73.34          \\ \hline
\multicolumn{1}{c|}{\multirow{4}{*}{\begin{tabular}[c]{@{}c@{}}Baseline + Context \\ Augmentation\end{tabular}}}                                & \multicolumn{1}{c|}{\textit{\begin{tabular}[c]{@{}c@{}}+ AGCA\\ w/o Gated Map\end{tabular}}}   & 71.33          & 45.35          & 74.32          \\
\multicolumn{1}{c|}{}                                                                                                                           & \multicolumn{1}{c|}{\textit{+ AGCA}}                                                          & 72.28          & 47.07          & 75.13          \\ \cline{2-5} 
\multicolumn{1}{c|}{}                                                                                                                           & \multicolumn{1}{c|}{\textit{+ DLCA}}                                                          & 71.25          & 46.72          & 74.59          \\ \cline{2-5} 
\multicolumn{1}{c|}{}                                                                                                                           & \multicolumn{1}{c|}{\textit{+ AGCA \& DLCA}}                                                  & \textbf{73.14} & \textbf{47.35} & \textbf{75.43} \\ \hline
\multicolumn{1}{c|}{\multirow{2}{*}{\begin{tabular}[c]{@{}c@{}}Baseline + Context\\ Augmentation + Consistent \\ Attention Weighting\end{tabular}}} & \multicolumn{1}{c|}{\textit{\begin{tabular}[c]{@{}c@{}}+ CA in  Optimization Only\end{tabular}}} & 74.94          & \textbf{50.67} & \textbf{77.03} \\
\multicolumn{1}{c|}{}                                                                                                                           & \multicolumn{1}{c|}{\textit{\begin{tabular}[c]{@{}c@{}}+ CA in Optimization \\ and Matching (MTLDesc)\end{tabular}}}     & \textbf{78.66} & 47.16          & 75.92          \\ \hline
\multicolumn{1}{c|}{\multirow{2}{*}{\begin{tabular}[c]{@{}c@{}}Current SOTA \\ Comparison Method\end{tabular}}}                                 & \multicolumn{1}{c|}{\textit{DISK} \textit{(2K)}}                                                       & 76.09          & 44.36          & 68.14          \\
\multicolumn{1}{c|}{}                                                                                                                           & \multicolumn{1}{c|}{\textit{DISK}\textit{(8K)}}                                                       & 77.58          & 45.18          & 69.81          \\ \hline
\end{tabular}}
}
\begin{tabular}{c}
    \\
\end{tabular}
\vspace{-0.5cm}
\caption{Ablation study on HPatches benchmark. 
We report 
metrics at a 3px error threshold for different variants of our MTLDesc.
\textbf{\emph{FP Keypoints}} means Local features Detection with Feature Pyramid.
\textbf{\emph{AGCA}} means Adaptive Global Context Augmented Module.
\textbf{\emph{DLCA}} means Diverse Local Context Augmented Module.
\textbf{\emph{CA}} means Consistent Attention. 
\vspace{-0.5cm}
}
\label{tab:module}
\end{table}

\noindent\textbf{Comprehensive Ablation Studies:}
\label{sec:ablation}
Ablation studies are reported in Tab.~\ref{tab:module}. 
Our proposed data construction method is used to re-implement the SuperPoint named $our \ impl.$ as a stronger baseline with higher \textit{MMA} and \textit{M.S.} scores compared to the SuperPoint $orig.$ 
After applying proposed \emph{FP Keypoints}, all metrics have been significantly improved, due to more accurate keypoints localization.
Almost all metrics get steady improvement after incrementally applying AGCA and DLCA. Note that the Gated Map in AGCA can further improve the performance, especially for M.S. score.
In addition, we also compared some related operations that can implicitly improve context awareness including DCN in~\cite{aslfeat}, Hypercolumns in~\cite{mlifeat}, UNet-like backbone in~\cite{tyszkiewicz2020disk} and ASPP in~\cite{aspp}. Our proposed Context Augmentation is superior to these alternative context augmentation operations.
Furthermore, all metrics still have a significant improvement with only using consistent attention in optimization ($CA\ in\ Optimization $) and not using consistent attention in matching. This indicates that our proposed Consistent Attention Weighted Triplet Loss can independently improve the performance of the descriptors substantially.
Different from the above settings, the consistent attention weighted descriptors ($CA\ in\ Matching $) are also used for matching to obtain an \textit{MMA} score of {\bf 78.66}.
Our MTLDesc significantly exceeds the current state-of-the-art DISK ~\cite{tyszkiewicz2020disk} on all metrics.
\vspace{-0.3cm}
\begin{figure}[H]
\begin{center}
  \includegraphics[width=1\linewidth]{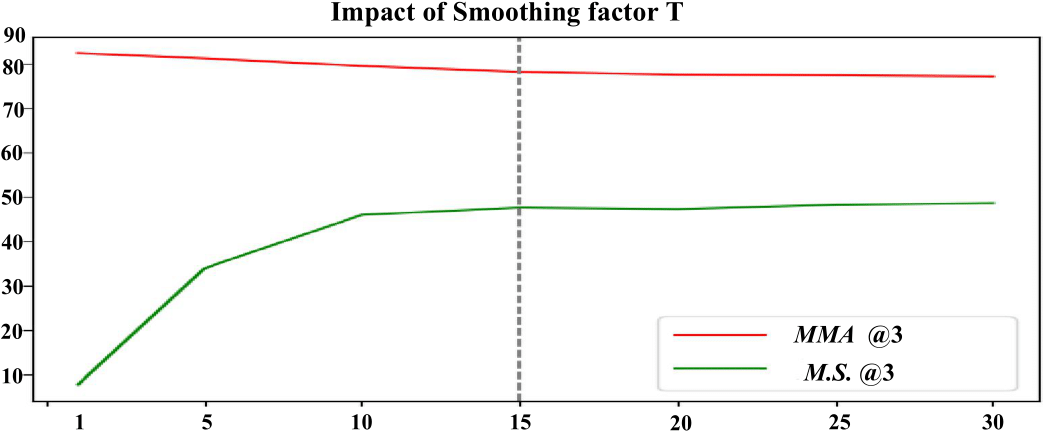}

\end{center}
   \caption{
 Impact of Smoothing factor. \textit{MMA@3} and \textit{M.S.@3} on the Hpatches evaluation under different \textit{T} settings.
   }
\label{fig:T}
\end{figure}

\vspace{-0.3cm} 

\begin{figure*}[htb]
\setlength{\abovecaptionskip}{-0.1cm} 
\setlength{\belowcaptionskip}{-0.5cm}
\begin{minipage}[H]{0.7\linewidth}
\centering
\includegraphics[width=5.4 in]{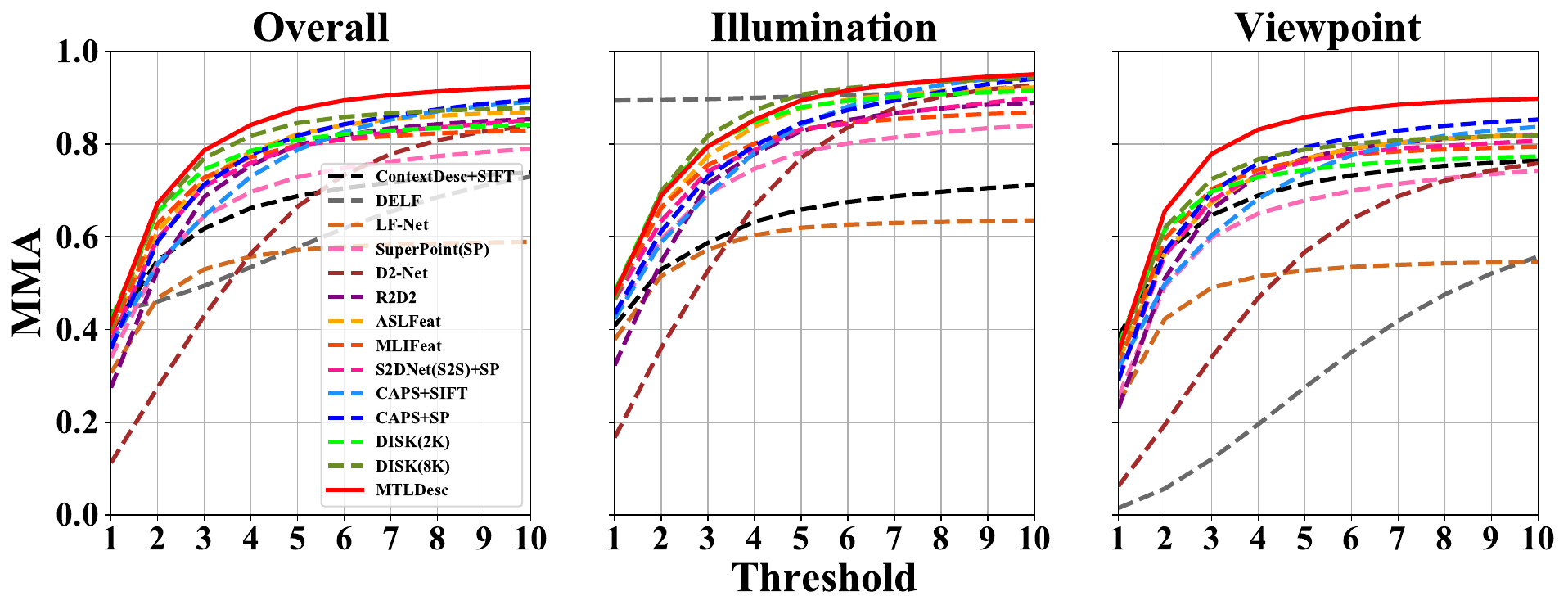}
\end{minipage}%
\begin{minipage}[H]{0.38\linewidth}
\centering %
\includegraphics[width=1.55 in]{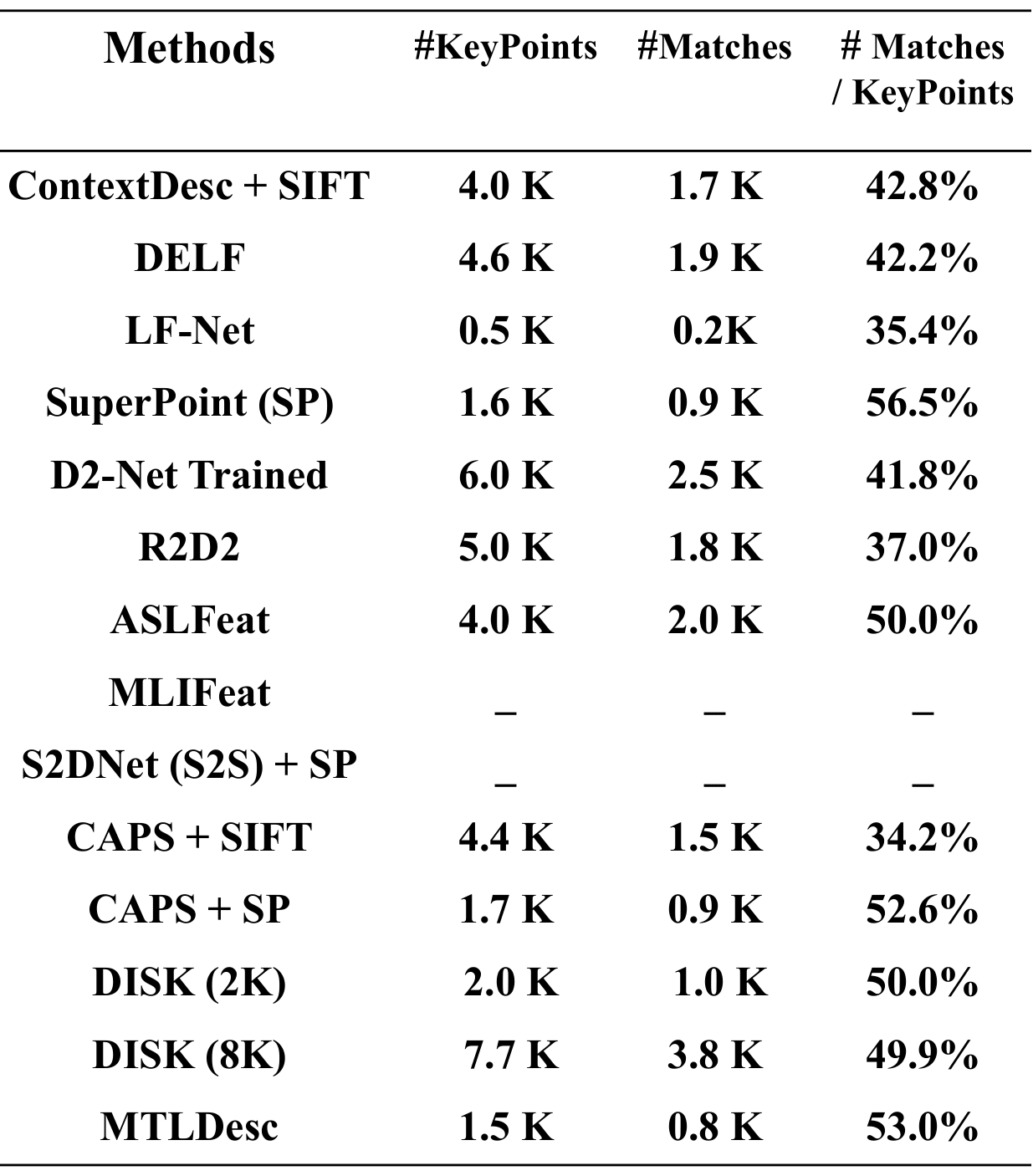}
\end{minipage}
 \caption{Left: Comparisons on HPatches with different thresholds Mean Matching Accuracy. MTLDesc achieves the best overall performance at all thresholds. Right: We also give the mean number of keypoints, nearest neighbour matches between keypoints, and the ratio between them. MTLDesc has a high utilization rate of keypoints.}

\label{fig:hpatch}
\end{figure*}

\noindent\textbf{Impact of Smoothing factor T:}
\label{sec:T}
The $T$ is used to adjust the effect of consistent attention on the loss function.
When $T$ becomes larger, the \emph{weight} $\frac{e^{^{\omega/T}}}{\sum_{i=1}^{N} e^{^{\omega_{i}/T}}}$ in Eq.~\ref{eq:attention triplet} is smoothed and the \emph{weights} of different samples are closer. 
On the contrary, the \emph{weights} 
will be concentrated on some better samples (easier to optimize) with a smaller $T$. 
It will cause the network to only optimize better samples and the optimization of other normal samples is undermined.
In the evaluation, these non-optimized normal samples lead to fewer possible matches (including correct and incorrect matches) and more correct matches in possible matches.
The increase in the proportion of correct matches in possible matches will lead to a higher \textit{MMA}. However, the reduced number of possible matches also contains some correct matches, which leads to a lower \textit{M.S.} score.
To obtain both accurate and dense local features matching results, it is necessary to balance \textit{M.S.} and \textit{MMA} by adjusting the value of parameter $T$.
As observed in Fig.~\ref{fig:T}, the best balance between \textit{M.S.} and \textit{MMA} is achieved when $T$ is set to 15.

\noindent \textbf{Comparisons with Advanced Local Descriptors:}
In Fig.~\ref{fig:hpatch}, we compare our MTLDesc with advanced local descriptors~\cite{lfnet,luo2019contextdesc,DELF,superpoint,d2net,r2d2,aslfeat,mlifeat,germain2020s2dnet,wang2020caps,tyszkiewicz2020disk} on  HPatches benchmark.
All methods use the optimal configuration and results reported in their papers, while MTLDesc notably outperforms these methods in all thresholds under overall \textit{MMA}.
Although the performance of the recent DISK is closest to our method, we should note that when DISK uses {\bf 2 K keypoints} (equivalent to our 1.5 K), the performance is much lower than our method. 
It is worth mentioning that DISK does not exceed our method even if using {\bf unfair 8 K keypoints}.

\begin{table}[t]
\centering
\setlength{\abovecaptionskip}{0cm} 
		\setlength{\belowcaptionskip}{-0.2cm}
\resizebox{\linewidth}{!}{
\renewcommand\arraystretch{1.18} {
\begin{tabular}{cccccc}
\hline
\multicolumn{6}{c}{\textbf{Aachen Day-Night v1.1 Benchmark}}                                                                                                                              \\ \hline
\multirow{2}{*}{\textbf{Method}} & \multirow{2}{*}{\textbf{Dim}} & \multirow{2}{*}{\textbf{Features}} & \multicolumn{3}{c}{\textbf{Correctly localized queries}}                          \\
                                 &                               &                                    & \textit{0.25m,2$^{\circ}$} & \textit{0.5m,5$^{\circ}$} & \textit{5m,10$^{\circ}$} \\ \hline
ROOT-SIFT                        & 128                           & 11 K                               & 53.4                       & 62.3                      & 72.3                     \\
DSP-SIFT                         & 128                           & 11 K                               & 40.3                       & 47.6                      & 51.3                     \\
SuperPoint                    & 256                           & 7 K                                & 68.1                       & 85.9                      & 94.8                     \\
D2Net                           & 512                           & 14 K                               & 67.0                       & 86.4                      & 97.4                     \\
R2D2                           & 128                           & 10 K                               & 70.7                       & 85.3                      & 96.9                     \\
ASLFeat                        & 128                           & 10 K                               & 71.2                       & 85.9                      & 96.9                     \\
CAPS + SuperPoint                  & 256                           & 7 K                                & 71.2                       & 86.4                      & \textbf{97.9}            \\
DISK                            & 128                           & 10 K                               & 72.8                       & 86.4                      & 97.4                     \\\hline
Our MTLDesc                         & 128                           & 7 K                                & \textbf{74.3}              & \textbf{86.9}             & 96.9                     \\ \hline
\end{tabular}}
}
\resizebox{\linewidth}{!}{
\renewcommand\arraystretch{1.18} {
\begin{tabular}{ccccccc}
\hline
\multicolumn{7}{c}{\textbf{InLoc Benchmark}} \\ \hline
\multicolumn{1}{c|}{\multirow{2}{*}{\textbf{{Method}}}} & \multicolumn{6}{c}{\textbf{{Localized queries}}\emph{(\%, 0.25m/0.5m/1.0m)}} \\ \cline{2-7} 
\multicolumn{1}{c|}{} & \multicolumn{3}{c|}{\textbf{{DUC1}}} & \multicolumn{3}{c}{\textbf{{DUC2}}} \\ \hline
\multicolumn{1}{c|}{SuperPoint} & 39.9 & 55.6 & \multicolumn{1}{c|}{67.2} & 37.4 & 57.3 & \textbf{70.2} \\
\multicolumn{1}{c|}{D2Net} & 39.9 & 57.6 & \multicolumn{1}{c|}{67.2} & 36.6 & 53.4 & 61.8 \\
\multicolumn{1}{c|}{R2D2} & 36.4 & 57.1 & \multicolumn{1}{c|}{\textbf{73.7}} & 44.3 & 60.3 & 68.7 \\
\multicolumn{1}{c|}{ASLFeat} & 36.4 & 56.1 & \multicolumn{1}{c|}{66.7} & 36.6 & 55.7 & 61.1 \\ 
\multicolumn{1}{c|}{CAPS + SuperPoint} & 32.8 & 53.0 & \multicolumn{1}{c|}{64.6} & 32.8 & {58.8} & 64.1 \\
\multicolumn{1}{c|}{DISK} & 38.9 & 59.1 & \multicolumn{1}{c|}{67.7} & 37.4 & 57.3 & 64.1 \\
\hline
\multicolumn{1}{c|}{Our MTLDesc} & \textbf{41.9} & \textbf{61.6} & \multicolumn{1}{c|}{72.2} & \textbf{45.0} & \textbf{61.1} & \textbf{70.2} \\ \hline
\end{tabular}}

}
\begin{tabular}{c}
    \\
\end{tabular}
\caption{Evaluation results of the Aachen Day-Night v1.1 and InLoc benchmarks. 
We report the percentage of successfully located images within three error thresholds. 
\vspace{-0.6cm}}
\label{tab:aachen}
\end{table}



\subsection{Comparisons on Visual Localization}
\noindent\textbf{Dataset and Evaluation:}
We resort to Aachen Day-Night visual localization  v1.1~\cite{aachen} and InLoc indoor visual localization~\cite{taira2018inloc} benchmarks to further demonstrate the effectiveness of our MTLDesc.
For a fair comparison, all methods use the same image matching pairs provided by the benchmarks and the same evaluation pipelines except for local features.
The number of maximum local features of all methods is limited to 20 K as reported in the previous methods.
For Aachen, our evaluation is performed via a localization pipeline based on COLMAP~\cite{colmap}.
For InLoc, we evaluate all methods based on HLOC~\cite{sarlin2019coarse_hloc}.
See supplementary material for more details.

\noindent\textbf{Results Analysis:}
The comparison results are shown in Tab~\ref{tab:aachen}.
Our MTLDesc evidently outperforms other local descriptors under the tolerance~($0.25m, 2^{\circ}$ and $0.5m, 5^{\circ}$) and achieves competitive performance under the tolerance~($5m, 10^{\circ}$ or $1m, 10^{\circ}$) on both Aachen outdoor and InLoc indoor benchmarks, validating the effectiveness of our MTLDesc for visual localization task, especially under high-precision requirements.
Our MTLDesc only deploys a relatively small training dataset~(1/40 of ASLFeat and does not include any indoor scenes data) and a short descriptor representation~(128 vs 256, 512) to achieve the state-of-the-art performance, revealing the great potential for further improvement.

\section{Conclusion}

In this work, we propose a novel method named \textbf{MTLDesc} to cope with the local features detection and description simultaneously.
In order to make our descriptor ``look wider to describe better'', the Context Augmentation and Consistent Attention Weighting is designed to give descriptors a context awareness beyond the local region, while the Local Features Detection with Feature Pyramid is presented to obtain accurate and reliable keypoints localization.
We have conducted thorough experiments on the standard HPatches, Aachen and InLoc benchmark, and validate our MTLDesc can achieve state-of-the-art performance among local descriptors.

\section{Acknowledgements}
This work was supported in part by the National Natural Science Foundation of China (Nos. $61620106003$, $61971418$, U$2003109$, $62171321$, $62071157$, $62162044$  and $61771026$) and in part by the Open Research Fund of Key Laboratory of Space Utilization, Chinese Academy of Sciences (No. LSU-KFJJ-2021-05), Open Research Projects of Zhejiang Lab (No. 2021KE0AB07), and this work was partially sponsored by CAAI-Huawei MindSpore Open Fund.

\bibliography{aaai22}
\end{document}